\documentclass[11pt]{article}

\usepackage[margin=1in]{geometry}
\usepackage{microtype}
\usepackage{graphicx}
\usepackage{subfigure}
\usepackage{booktabs} 
\usepackage{amsmath,amssymb}
\usepackage{color}
\usepackage{hyperref}
\usepackage{float}

\title{Revisiting SVD and Wavelet Difference Reduction\\
for Lossy Image Compression: A Reproducibility Study}

\author{
  Alena Makarova\\
  Oregon State University\\
  \texttt{makarova@oregonstate.edu}
}

\date{} 

\begin{document}
\maketitle

\begin{abstract}
This work presents an independent reproducibility study of a lossy image
compression technique that integrates singular value decomposition (SVD) and
wavelet difference reduction (WDR). The original paper claims that combining
SVD and WDR yields better visual quality and higher compression ratios than
JPEG2000 and standalone WDR. I re-implemented the proposed method, carefully
examined missing implementation details, and replicated the original
experiments as closely as possible. I then conducted additional experiments on
new images and evaluated performance using PSNR and SSIM. In contrast to the
original claims, my results indicate that the SVD+WDR technique generally does
\emph{not} surpass JPEG2000 or WDR in terms of PSNR, and only partially
improves SSIM relative to JPEG2000. The study highlights ambiguities in the
original description (e.g., quantization and threshold initialization) and
illustrates how such gaps can significantly impact reproducibility and
reported performance.
\end{abstract}

\section{Introduction}

High-resolution satellite imagery plays a central role in applications such as glacier monitoring, debris-cover detection, and large-scale environmental analysis. However, the substantial size of such imagery imposes significant constraints on storage, transmission, and computational efficiency, particularly when used in machine learning workflows involving convolutional neural networks. These challenges naturally motivate the exploration of effective lossy compression techniques that can reduce data volume while preserving essential visual and structural information.

In this context, I examined a published image-compression method that integrates singular value decomposition (SVD) with wavelet difference reduction (WDR). Both techniques are well-established: SVD provides a compact low-rank representation that retains dominant image features, whereas WDR exploits multiresolution structure to achieve high compression ratios. The original paper claims that combining SVD and WDR yields improved visual quality relative to JPEG2000 and standalone WDR. Given the potential relevance of such a method for large scientific imagery, these claims warrant careful evaluation.

This study presents a detailed reproducibility analysis of the proposed SVD+WDR technique. I re-implemented the method based solely on the descriptions provided in the original publication, documented ambiguities in the algorithmic specification, and reproduced the reported experiments where possible. I also conducted additional tests on new datasets, including imagery relevant to glacier-segmentation research, to assess the technique’s robustness across diverse content.

The goal of this work is twofold: (1) to evaluate whether the reported performance advantages hold under a transparent, independently implemented pipeline, and (2) to highlight methodological details that meaningfully affect reproducibility in image-compression research. The results show substantial deviations from the original claims, underscoring the importance of clear algorithmic descriptions and thorough validation across varied image domains.

\section{Overview of SVD and WDR}

The background of the techniques used in the paper, singular value decomposition (SVD) and wavelet difference reduction (WDR), lies in their respective approaches to image compression.

SVD is a well-established technique in linear algebra as it decomposes a matrix into singular values, left singular vectors, and right singular vectors, providing a compact representation of the matrix. By discarding less significant components with smaller singular values, SVD enables matrix approximation with reduced dimensions. In image compression, SVD represents an image matrix with a lower rank approximation, allowing for a substantial reduction in file size while preserving crucial visual information.

On the other hand, WDR is a compression technique based on the wavelet difference algorithm. Wavelet difference reduction focuses on encoding the differences between wavelet coefficients at different scales, resulting in efficient data representation. This technique exploits the spatial frequency information in an image and achieves high compression ratios. However, it may introduce some level of visual degradation due to the lossy nature of the compression.

Image compression is crucial in machine learning, particularly in training deep learning models for image recognition. Consider a scenario with a large dataset of images consisting of thousands or even millions of high-resolution pictures. With compression, these images' storage and processing requirements would be manageable.

Compression techniques like the one discussed in this paper, which combines SVD and WDR, help solve this problem. The image is first compressed using SVD, providing a high-quality image by capturing essential features. Then, the compressed image is further compressed using WDR to achieve the desired compression ratio. By combining these techniques, the authors aim \cite{Orig} to balance image quality and compression ratio, taking advantage of the high-quality representation provided by SVD and the high compression ratios offered by WDR.

The proposed image compression technique in the paper is related to matrix analysis through the use of singular value decomposition (SVD). 

We can represent any image as a matrix \textbf{A} of numbers with elements representing the pixels. 
\begin{center}
$\textbf{A}_{m \times n} = \textbf{U}_{m \times m} \pmb{\Sigma}_{m \times n} (\textbf{V}_{n \times n})^T,$
\end{center}

Where \textbf{U} and \textbf{V} are orthogonal matrices, and $\pmb\Sigma$ is  a diagonal matrix with singular values in descending order. 
Compression is achieved by reducing the rank of the input matrix by eliminating small singular values $\sigma_i$. This decomposition allows for a more compact image representation, resulting in efficient compression.
\[
\pmb\Sigma_{m \times n} = 
\begin{bmatrix}
\overline{\pmb\Sigma}_{p \times q} & 0 \\

0 & \ddots
\end{bmatrix} p \leq m \text{ and } q\leq n
\] 

The number of non-zero elements on the diagonal of $\pmb\Sigma$ determines the matrix rank. A smaller rank $\pmb\Sigma$ is used to approximate the original matrix, obtained by disregarding low, singular values. 
\begin{center}
$\textbf{U}_{m \times m} = [\tilde{\textbf{U}}_{m \times p} \text{ }\tilde{\textbf{U}}_{m \times (m-p)}]$

$\textbf{V}_{n \times n} = [\tilde{\textbf{V}}_{n \times q} \text{ }\tilde{\textbf{V}}_{n \times (n-q)}]$
\end{center}
The reconstructed image is obtained by multiplying the reduced matrices  \textbf{U} and  \textbf{V} with the modified $\pmb\Sigma$. 

\begin{center}
$\textbf{A}_{m \times n} = \overline{\textbf{U}}_{m \times p} \overline{\pmb{\Sigma}}_{p \times q} (\overline{\textbf{V}}_{n \times q})^T,$
\end{center}

This compression process is lossy since the ignored singular values cannot be recovered.

The suitability of SVD for compression lies in its ability to reconstruct an image using fewer singular values. This property allows for efficient image compression, as demonstrated in the reconstruction of Lena's picture with different amounts of singular values (see the code). By effectively compressing the image through SVD, the visual quality is preserved while achieving compression.

The technique I am going to describe solves several challenges in image compression. 
\begin{itemize}
  \item Balancing image quality and compression ratio.
  The technique balances having high-quality images and achieving high compression ratios. It combines two methods, SVD and WDR, to maximize their strengths. SVD provides good image quality but lower compression ratios, while WDR offers higher compression ratios but potentially lower image quality. The proposed technique leverages the advantages of both methods to provide high-quality images at desired compression ratios.
  \item Reducing storage and transmission costs. The technique effectively compresses images, resulting in smaller file sizes. This reduction in size leads to cost savings in terms of storage space and bandwidth required for transmitting the images. 
  \item Outperforming existing techniques. The proposed method is compared to other state-of-the-art techniques like JPEG2000 and WDR. The experimental results show that it performs better regarding image quality, as measured by the peak signal-to-noise ratio (PSNR). Experimental results demonstrate that the proposed technique outperforms existing techniques regarding image quality. It offers better visual quality at higher compression ratios compared to other methods.
\end{itemize}

\section{Method}

\begin{figure}[H]
\vskip 0.2in
\begin{center}
\centerline{\includegraphics[width=0.6\columnwidth]{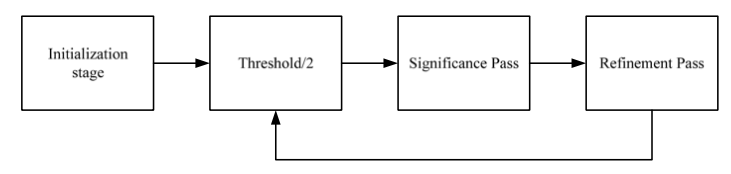}}
\caption{Block diagram of Bit-Plane encoding used by WDR.}
\label{11}
\end{center}
\vskip -0.2in
\end{figure}
\begin{figure}[H]
\vskip 0.2in
\begin{center}
\centerline{\includegraphics[width=0.6\columnwidth]{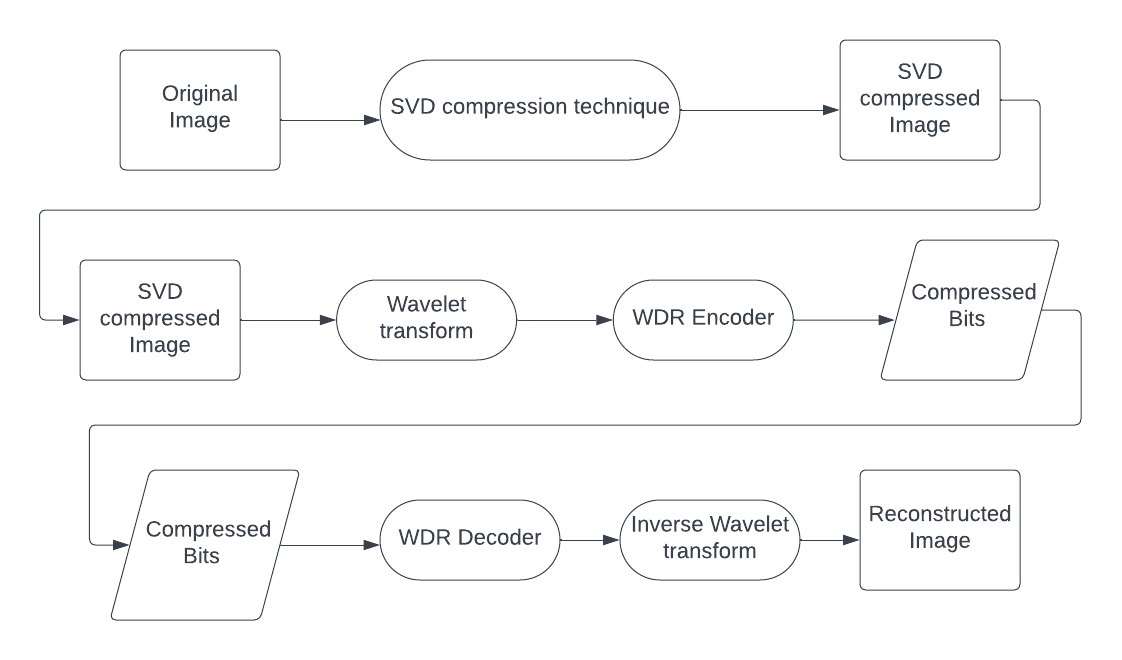}}
\caption{Block diagram of the proposed image compression technique.}
\label{block}
\end{center}
\vskip -0.2in
\end{figure}

During the paper analysis, it was considered to divide the proposed method  \cite{Method} into small steps that could be observed on Figure \ref{block}. 
\subsection {SVD}
First, we should apply SVD to the image in the proposed method. The SVD decomposition produces matrices \textbf{U}, $\pmb\Sigma$, and \textbf{V}, where $\pmb\Sigma$ contains the singular values representing each component's importance.

In the proposed SVD method, we take an image and a parameter $k$ that determines the number of singular values used for the approximation. Then we can calculate the SVD of the image using $np.linalg.svd $ function, and perform the approximation by multiplying the selected components.

\begin{center}
$\textbf{A}_{m \times n} = \overline{\textbf{U}}_{m \times k} \overline{\pmb{\Sigma}}_{k \times k} (\overline{\textbf{V}}_{n \times k})^T,$
\end{center}

It's essential to consider the trade-off between the approximation quality and the compression rate to analyze the method. Increasing the value of $k$ will result in a higher quality approximation as more critical components are included. However, it will also require more data to encode the approximation, leading to a lower compression rate. On the other hand, reducing the value of $k$ will increase the compression rate but may result in a lower-quality approximation.

\begin{figure}[H]
\vskip 0.2in
\begin{center}
\centerline{\includegraphics[width=0.6\columnwidth]{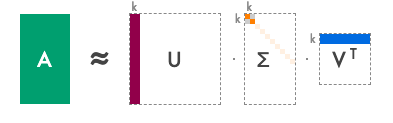}}
\caption{SVD}
\label{svd}
\end{center}
\vskip -0.2in
\end{figure}

To compare the results with different compression ratios (CR), we can compute them based on the dimensions of the image and the chosen value of $k$ (Figure ~\ref{svd}). 

\begin{center}
$CR = \text{US} / \text{CS}$
\end{center}

where US (UNCOMPRESSED SIZE) proportional to number of pixels, and  CS (COMPRESSED SIZE)
approximately proportional to

\begin{center}
CS = $m \times k + k + k \times n = k \times (1 + m + n)$

\end{center}

The higher the compression ratio, the more significant the compression.

\subsection{WDR}

The Wavelet Difference Reduction (WDR) method is a compression technique that can be used in both lossy and lossless scenarios. In our case, the lossy version of WDR is employed and it means that we can't recover the image after compression. The process involves applying a wavelet transform (DWT) to the image, followed by bit plane encoding of the transform values on Figure \ref{11}. 

\subsubsection{Wavelet Transform}

First, I performed a wavelet-based compression. As a result, an image is decomposed into two sets of coefficients: the approximation coefficients and the detail coefficients. Approximation coefficients represent the low-frequency components or the overall structure of the image, and detail coefficients represent the high-frequency components or the finer details of the image. Since the authors didn't mention if they used approximation coefficients, I decided to discard them and use only detailed coefficients for simplicity, which I merged into a single array. 
\subsubsection{WDR Encoding}
\textbf{Initialisation stage}

In the initialization stage of WDR compression, we choose a special number called the initial threshold ($T_0$). This threshold is bigger than any of the numbers we're working with. We make sure that at least one of our numbers has a magnitude (size) equal to half of the initial threshold value. This helps us establish a starting point for the compression process.
According to the authors' instructions, it is recommended to select the threshold value so that at least one number from the list is equal to half of the threshold. However, considering the iterative nature of the convergence process, it is necessary to perform multiple loops rather than just one. In this case, the condition of having one or more numbers equal to half the threshold becomes more applicable.  So that,
I decided to take the threshold value that what be equal to the value of the maximum element in the list by 2. Thus, I can be sure that my code will execute at least once. 

\textbf{Threshold$/$2}

After the initialization stage, we update the threshold value. The new threshold (T) is obtained by dividing the previous threshold ($T_{k-1}$) by 2. So, if my maximum value in a list were, for example, 64, the new value of $T$ would be 32. 

\textbf{Significance pass}

To perform the next step, we apply the difference reduction method. The main focus is on the differences between indexes rather than the absolute ones. These differences provide a compact representation of the sequence of significant values. Using the differences, we can reconstruct the original sequence by starting from a known initial index and adding the differences successively.

Let's say we have a threshold value ($T$) set to 32. We examine the transformed values ($w$) and identify the ones that fall within the range 
$$ T \leq |w(i)| \leq 2T$$ defined by the threshold. 
In this case, we have w(1) = 68, w(2) = -32, w(5) = -75, and w(36) = 108.

We record the indexes of these significant values, which are 1, 2, 5, and 36, respectively.

\textbf{Difference Reduction}
\begin{itemize}
\item
We calculate the differences between consecutive indexes of the significant values we found. For example, the difference between index 2 and index 1 is 1, the difference between index 5 and index 2 is 3, and so on.
\item
We convert these differences into binary form. For example, the difference 1 is represented as $(1)_2$, the difference 3 is represented as $(11)_2$, and the difference 31 is represented as $(11111)_s2$.
\item
We create a symbol stream using these binary representations and the signs of the significant values. The symbols '+' and '-' are used to indicate the signs, and the binary digits represent the steps needed to reach the next index. 
Also, in the binary expansions of the differences, the most significant bit is always 1. This bit is consistent and does not provide new information. Therefore, it can be dropped or ignored during the compression process.
Instead of including the dropped most significant bit, we utilize the signs of the significant transform values as separators in the symbol stream. 
In this case, the symbol stream is 
$$+ - +1 +1111$$
\end{itemize}

\textbf{Refinement Pass}

In the Refinement Pass, we aim to reduce the error or discrepancy between the original transform values and their quantized versions.

We start with a threshold value of 32.

According to the information presented in the paper, a specific procedure was outlined for handling significant values. It was indicated that each significant value should be examined to determine whether its absolute magnitude is equal to or greater than twice the threshold value (2T). In such cases, a modification was suggested, wherein the value would be rounded to the nearest threshold multiple. This step resembles a quantization process, wherein the value is discretized or approximated to a specific level.

However, the paper did not provide comprehensive details or explanations regarding the quantization step, leaving certain aspects ambiguous. Consequently, it is unclear how the quantization process was precisely implemented and what specific criteria or methodologies were employed. Further clarification and additional information from the authors would be valuable to understand better the quantization procedure and its implications within the context of the proposed method.

In our example:
\begin{itemize}
    \item The first significant value, w(1) = 68, satisfies the condition $ |w(1)| \geq 2T = 68 \geq 64$, so we round it to 64 as the closest interval is $[64, 96)$.
    \item The second significant value, w(2) = -32, doesn't satisfy our condition $ |w(2)| \geq 2T$, so we leave the value.
    \item The third significant value, w(5) = -75, satisfies the condition $ |w(5)| \geq 2T = |75|\geq 64$, so we round it to -64 as the closest interval is $[64, 96).$
    \item The fourth significant value, w(36) = 108, also satisfies the condition $ |w(36)| \geq 2T = 108 \geq 64$, so we round it to 96 as the closest interval is $[96, 128).$
\end{itemize}

The refined values obtained after the refinement pass are 64, -32, -64, 96. These refined values provide better approximations of the original transform values and help reduce the overall error in the compression process.

The loop on Figure 1 is needed as the WDR compression process involves dividing the image into parts called bit planes. Each bit plane represents a specific level of detail in the image.

To compress the image, I went through each bit plane individually. I started with the most important bit plane, which contains the crucial details of the image. Then, I moved to the next bit plane, which contains less important details, and so on.

I performed a series of operations for each bit plane to encode the information in that plane. This encoding process helps reduce the data needed to represent the image while preserving its essential features.

In their study, the authors reported conducting 12 cycles to achieve the optimal outcome in their experimental setup. However, in my attempt to replicate their process, I could only perform nine cycles. This discrepancy may be attributed to the initial value assigned to the threshold parameter. Unfortunately, the authors did not provide specific details or values regarding threshold initialization, which is a crucial aspect of the method. It would greatly benefit future readers and researchers if the authors could include this essential information, as it plays a significant role in reproducing and understanding their results accurately.

\subsubsection{Compression Rate for WDR}

I calculate the compression rate by comparing the sizes of the original data and the compressed data. 

\begin{center}
$CR_{WDR} = OR / CZ$
\end{center}
where 
\begin{itemize}
    \item $CR_{WDR}$ the compression rate of WDR
    \item $OR$ the size of the original data before compression
    \item $CZ$ the size of the compressed data (refined values)
\end{itemize}    

\subsubsection{Reverse WDR and Wavelet Transform}

In order to decompress the image that was previously encoded using the Wavelet Difference Reduction (WDR) technique, a decoding process was performed. The decoding involved reshaping the sequence of refined values obtained during the compression stage into the appropriate array format. This reshaping step ensured that the refined values could be accurately inserted into the Inverse Wavelet Transform operation. The lossy uncompressed image was obtained by applying the Inverse Wavelet Transform to the reshaped array, effectively reversing the compression process and restoring the image to its original form.

\subsection{Overall Compression Rate}

The paper's authors described a method that combines the compression rates of singular value decomposition (SVD) and Wavelet Difference Reduction (WDR) to determine the overall compression rate (multiplying these two values to get a compression rate for a novel method). However, the paper does not provide explicit formulas or guidelines for calculating SVD and WDR compression rates individually. As a result, it is unclear how these compression rates are determined, making it difficult to understand the rationale behind multiplying these two values.

\section{Experiment}
\subsection{Repeating the experiments in the paper}

The implementation used in this reproducibility study will be released at:
\url{https://github.com/AlenaMak1995/svd-wdr-compression}.

In the paper, the authors mentioned the usage of pictures depicting Airfield and Boats, along with the well-known images of Goldhill and Lena. However, I could only locate the images of Goldhill, Lena, and Peppers. Since the specific content of the Airfield and Boats images was not provided, I made the decision to select alternative pictures that are potentially similar to the ones used by the authors.

This approach ensures that the visual examples I utilized align with the intended context of the original images mentioned in the paper. Although there may be slight variations between the selected images and the actual Airfield and Boats pictures, the chosen alternatives aim to provide a reasonable representation within the scope of the authors' work.

To evaluate and compare the results, the metrics like PSNR and SSIM are used in the paper.

\subsubsection{Mean Square Error (MSE)}

MSE is the average of the squared differences between the pixel values of the original image and the compressed image. It measures the overall difference in quality between the two images.

The mathematical representation of MSE is given by:

\begin{equation*}
MSE = \frac{1}{mn} \sum_{y=1}^{n} \sum_{x=1}^{m} (f_A(x,y) - f_{Ak}(x,y))^2
\end{equation*}

where $m$ and $n$ are the dimensions of the images, $f_A(x,y)$ represents the pixel value of the original image, and $f_{Ak}(x,y)$ represents the pixel value of the compressed image.

\begin{figure}[H]
\begin{center}
\centerline{\includegraphics[width=0.6\columnwidth]{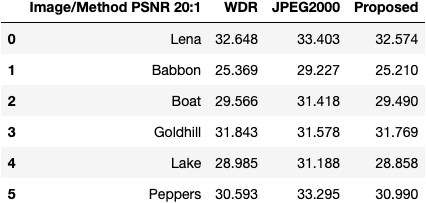}}
\caption{PSNR values (dB) for 20:1 compression.}
\label{1}
\end{center}
\end{figure}

\begin{figure}[H]
\begin{center}

\centerline{\includegraphics[width=0.6\columnwidth]{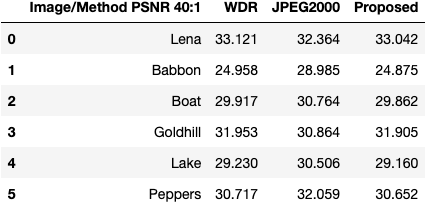}}
\caption{PSNR values in dB for 40:1 compression. }
\label{2}
\end{center}
\vskip -0.1in
\end{figure}

\begin{figure}[H]

\begin{center}

\centerline{\includegraphics[width=0.6\columnwidth]{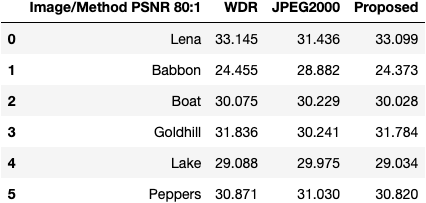}}
\caption{PSNR values in dB for 80:1 compression.}
\label{3}
\end{center}
\vskip -0.2in
\end{figure}
\subsubsection{Peak Signal-to-Noise Ratio (PSNR)}

PSNR is a measure of the quality of a compressed image in terms of the ratio of the maximum possible power of a signal to the power of corrupting noise. It is often expressed in terms of the logarithmic decibel scale.

The PSNR (in dB) is defined mathematically as:

\begin{equation*}
PSNR = 10 \log_{10} \left( \frac{MAX_i^2}{MSE} \right)
\end{equation*}

where $MAX_i$ is the maximum possible pixel value 

\begin{figure}[H]
\begin{center}
\centerline{\includegraphics[width=0.6\columnwidth]{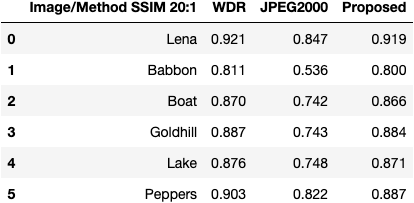}}
\caption{SSIM values in dB for 20:1 compression. }
\label{4}
\end{center}
\end{figure}

\begin{figure}[H]

\begin{center}

\centerline{\includegraphics[width=0.6\columnwidth]{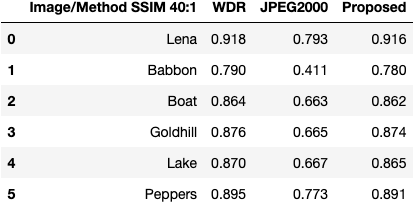}}
\caption{SSIM values in dB for 40:1 compression.}
\label{5}
\end{center}
\vskip -0.2in
\end{figure}

\begin{figure}[H]

\begin{center}

\centerline{\includegraphics[width=0.6\columnwidth]{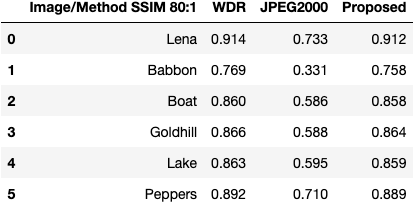}}
\caption{SSIM values in dB for 80:1 compression.}
\label{6}
\end{center}
\vskip -0.2in
\end{figure}
(e.g., 255 for 8-bit grayscale images) and $MSE$ is the Mean Square Error calculated between the original and compressed images.

\subsubsection{Structural Similarity Index (SSIM)}

SSIM is a metric that measures the structural similarity between two images. It considers not only the pixel-wise differences but also the structural information of the images.

The mathematical representation of SSIM is given by:

\begin{equation*}
SSIM(x, y) = \frac{{2 \mu_x \mu_y + c_1}}{{\mu_x^2 + \mu_y^2 + c_1}} \cdot \frac{{2 \sigma_{xy} + c_2}}{{\sigma_x^2 + \sigma_y^2 + c_2}}
\end{equation*}

where $x$ and $y$ are the input images, $\mu_x$ and $\mu_y$ are the means of $x$ and $y$ respectively, $\sigma_x$ and $\sigma_y$ are the standard deviations of $x$ and $y$ respectively, $\sigma_{xy}$ is the covariance of $x$ and $y$, and $c_1$ and $c_2$ are constants to stabilize the division.

The overall SSIM index is the average of SSIM values calculated over all the windows in the images.

Figure~\ref{1} presents a comprehensive comparison between the proposed techniques and JPEG2000 and WDR, utilizing the Peak Signal-to-Noise Ratio (PSNR) metric for a compression ratio of 20:1. To evaluate the performance of the proposed image compression technique across different compression ratios, Figure~\ref{2} and Figure~\ref{3} are provided, showcasing compression ratios of 40:1 and 80:1, respectively.
The images used in the experiments are all 512×512, 8-bit grayscale test images.

The PSNR values used for the JPEG2000 baseline were taken from the results
reported in \cite{10}. However, after reviewing \cite{10} more closely, it
became clear that the authors did not specify the exact test images used to
produce their JPEG2000 results. Moreover, the dataset described in \cite{10}
does not fully align with the images referenced in the paper under review.
Because of this mismatch, the origin of the JPEG2000 values reported by the
authors remains uncertain, making it difficult to verify their comparability
with the results obtained in this study.

After developing my own program for JPEG2000, it became evident that the achieved results were lower than initially anticipated. The obtained values did not align with the expected outcomes based on the comparison provided in the paper.

The comparative analysis presented in Figure~\ref{1}, ~\ref{2}, and ~\ref{3} reveals that the performance of the proposed image compression technique falls short of surpassing both JPEG2000 and WDR methods. 
Notably, the proposed technique generally does not outperform the WDR method in most instances. The authors of the paper mentioned discarding certain singular values in the SVD method, yet they did not specify the exact quantity discarded. It would greatly aid in comprehending the extent to which the proportion of discarded values from the SVD impacts the overall quality achieved by the proposed method. 

Figures ~\ref{4}, ~\ref{5}, and ~\ref{6} compare the SSIM values for the proposed technique with those of WDR and JPEG2000. The SSIM values show that the proposed technique outperforms the JPEG2000 technique only, but not the WDR.

\begin{figure}[H]

\begin{center}

\centerline{\includegraphics[width=0.6\columnwidth]{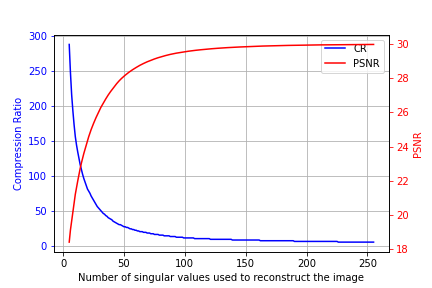}}
\caption{Graph showing PSNR, compression ratio and number of singular values used for peppers image.}
\label{exp1}

\end{center}
\vskip -0.2in
\end{figure}

Figure \ref{exp1} illustrates the relationship between the Peak Signal-to-Noise Ratio (PSNR) values in decibels (dB) and the compression ratio when varying the number of selected singular values in the proposed image compression technique. The experiment was conducted on a peppers image with 512 × 512 pixels dimensions. While the trend observed in the plot aligns with the findings reported in the referenced paper, it is important to note that the PSNR values obtained in our experiment are significantly lower than those reported in the paper.

\subsection{Conducting new experiments}

In order to conduct further experiments and broaden the evaluation of the proposed image compression technique, a new dataset consisting of different images was selected. The images in this dataset have a standardized size of 256 × 256 pixels and are represented in 8-bit grayscale format. Additionally, two well-known benchmark images, namely Lena and Goldhill, were included in the dataset for comparison purposes.

To assess the quality of the compressed images obtained from this new dataset, two  evaluation metrics - PSNR and SSIM - were employed.

\begin{figure}[H]
\begin{center}
\centerline{\includegraphics[width=0.6\columnwidth]{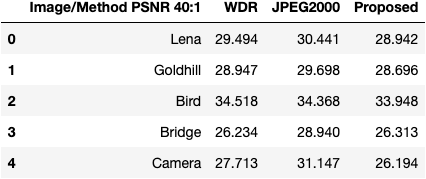}}
\caption{PSNR values in dB for 40:1 compression. }
\label{hill}
\end{center}
\end{figure}

Figures ~\ref{hill} and ~\ref{camera} provide an extensive comparative analysis between the proposed techniques and two established compression methods, JPEG2000 and WDR. The evaluation is conducted using two widely-used metrics, namely PSNR and SSIM, focusing on a compression ratio of 40:1.

The results obtained from Figures ~\ref{hill} and ~\ref{camera} highlight the performance of the different techniques in terms of their ability to preserve image quality during compression. JPEG2000 consistently exhibits superior performance in terms of PSNR across all tested scenarios, indicating its proficiency in maintaining high fidelity between the original and compressed images. However, when considering the SSIM metric, the proposed techniques demonstrate more favorable outcomes.

\begin{figure}[H]
\begin{center}

\centerline{\includegraphics[width=0.6\columnwidth]{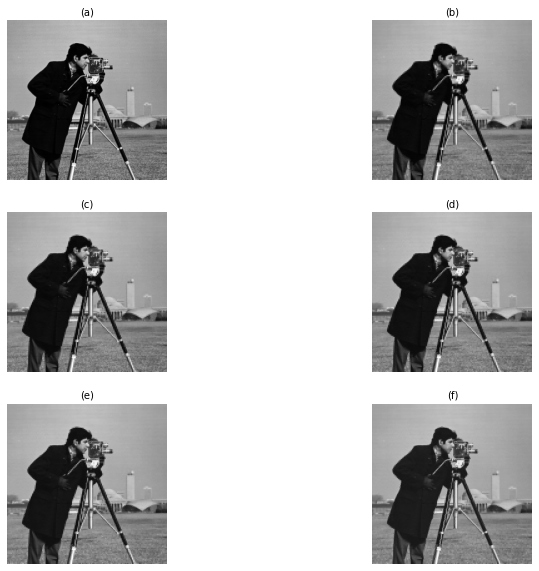}}

\caption{(a) Original uncompressed Camera image, (b) Camera image at compression ratio of 20:1, (c) Camera image at compression ratio of 40:1, (d) Camera image at compression ratio of 60:1, (e) Camera image at compression ratio of 80:1, (f) Camera image at compression ratio of 160:1}
\label{man}
\end{center}
\vskip -0.1in
\end{figure}

Figure ~\ref{man} showcases the visual quality of the compressed peppers image using different compression ratios achieved by the proposed technique. The figure compares the original pepper image and its compressed counterparts at compression ratios of 20:1, 40:1, 60:1, 80:1, and 160:1.

Through visual inspection, it can be observed that the compressed images maintain a high level of visual fidelity and resemblance to the original image, even at higher compression ratios. However, as the compression ratio increases, a slight degradation in visual quality becomes noticeable. At the 160:1 compression ratio, some disturbing noise artifacts emerge due to the loss of more significant singular values during the SVD-based compression process.
 
\begin{figure}[H]
\begin{center}

\centerline{\includegraphics[width=0.6\columnwidth]{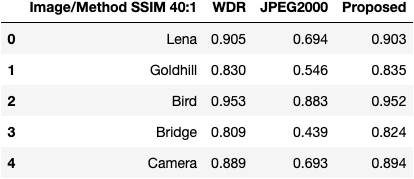}}
\caption{SSIM values in dB for 40:1 compression.}
\label{camera}
\end{center}
\vskip -0.1in
\end{figure}

\section{Conclusion}
In this reproducibility analysis, a novel lossy image compression technique that combines SVD and WDR was explored. The proposed approach was anticipated to surpass the performance of existing methods such as JPEG2000 and WDR. However, the experimental results obtained during the evaluation phase yielded different outcomes.

One significant challenge encountered during the paper review process was the lack of coherence and relevance among the referenced papers \cite{8}, \cite{9}. Many of the cited sources did not directly contribute to the subject matter, and attempting to obtain specific details often led to repetitive and redundant information. Additionally, crucial aspects of the proposed technique, such as the quantization step in the refinement pass and the selection of singular values, were not adequately addressed, leaving crucial gaps in understanding.

The performance comparison with JPEG2000 also presented certain ambiguities and complexities. Understanding the underlying reasons for the observed discrepancies required further investigation and analysis.

In conclusion, while the initial evaluation of the combined SVD and WDR compression technique did not yield the expected outcomes, the research journey has provided valuable insights and highlighted areas for improvement.

\bibliography{references}
\bibliographystyle{plain}

\appendix

\end{document}